\documentclass[wcp]{jmlr}

\usepackage{longtable}

\usepackage{booktabs}

\usepackage{hyperref}       
\usepackage{url}            
\usepackage{booktabs}       
\usepackage{amsfonts}       
\usepackage{nicefrac}       
\usepackage{microtype}      
\usepackage{xcolor}         
\usepackage{latexsym}
\usepackage{amssymb, amsmath}
\usepackage{enumitem}
\usepackage{graphicx}
\usepackage{multirow}
\usepackage{lineno}

\makeatletter
\let\Ginclude@graphics\@org@Ginclude@graphics 
\makeatother

\jmlrvolume{304,}
\jmlryear{2025}
\jmlrworkshop{ACML 2025}

\title{HIPPD: Brain-Inspired Hierarchical Information Processing for Personality Detection}

 \author{%
  \Name{Guanming Chen} \Email{gc3n21@soton.ac.uk}\\
  \Name{Lingzhi Shen} \Email{l.shen@soton.ac.uk}\\
  \Name{Xiaohao Cai} \Email{x.cai@soton.ac.uk}\\
  \Name{Shoaib Jameel} \Email{M.S.Jameel@southampton.ac.uk}\\
  \addr School of Electronics and Computer Science, University of Southampton, Southampton, United Kingdom
  \AND
  \Name{Imran Razzak} \Email{imran.razzak@mbzuai.ac.ae}\\
  \addr Department of Computational Biology, Mohamed bin Zayed University of Artificial Intelligence, Abu Dhabi, United Arab Emirates
}

\editors{Hung-yi Lee and Tongliang Liu}

\begin{document}

\makeatletter
\let \@jmlrpages \@empty
\makeatother

\maketitle

\begin{abstract}
Personality detection from text aims to infer an individual’s personality traits based on linguistic patterns. However, existing machine learning approaches often struggle to capture contextual information spanning multiple posts and tend to fall short in extracting representative and robust features in semantically sparse environments. This paper presents HIPPD, a brain-inspired framework for personality detection that emulates the hierarchical information processing of the human brain. HIPPD utilises a large language model to simulate the cerebral cortex, enabling global semantic reasoning and deep feature abstraction. A dynamic memory module, modelled after the prefrontal cortex, performs adaptive gating and selective retention of critical features, with all adjustments driven by dopaminergic prediction error feedback. Subsequently, a set of specialised lightweight models, emulating the basal ganglia, are dynamically routed via a strict winner-takes-all mechanism to capture the personality-related patterns they are most proficient at recognising. Extensive experiments on the Kaggle and Pandora datasets demonstrate that HIPPD consistently outperforms state-of-the-art baselines.
\end{abstract}


\section{Introduction}
Personality can be likened to a fingerprint: distinct to each individual, yet displaying common patterns. Although various theories on personality exist, they commonly concur that it is shaped by a complex mix of genetic, environmental, and experiential factors \citep{kozlova2024influence}. Among these, the Myers-Briggs Type Indicator (MBTI) \citep{myers1962myers} is a widely used and effective personality assessment tool. It categorises individuals into 16 different personality types based on four dimensions: Introversion vs. Extraversion (I/E), Sensing vs. Intuition (S/N), Thinking vs. Feeling (T/F), and Judging vs. Perceiving (P/J). 

Recent detection approaches, such as EERPD \citep{li2024eerpd}, integrate emotion regulation with emotional features, leveraging few-shot examples and chain reasoning. However, they may have high requirements for the quality and diversity of emotional data. The PsyAttention model \citep{zhang2023psyattention} improves personality detection accuracy by effectively encoding psychological features and reducing the number of features. However, during feature selection, some details crucial for specific application scenarios may be lost, potentially affecting the model's adaptability across different contexts. In \cite{hu2024llm}, the authors propose a method that utilises large language models (LLMs) to generate augmented textual data and explanations of personality labels from raw posts, focusing on semantic, emotional, and linguistic aspects. While AI-generated data can provide rich information, it may also introduce more severe challenges \citep{shen2025gamed}, such as causing hallucinations and amplifying biases.

\begin{figure}[htbp]
    \centering
    \includegraphics[scale=0.09]{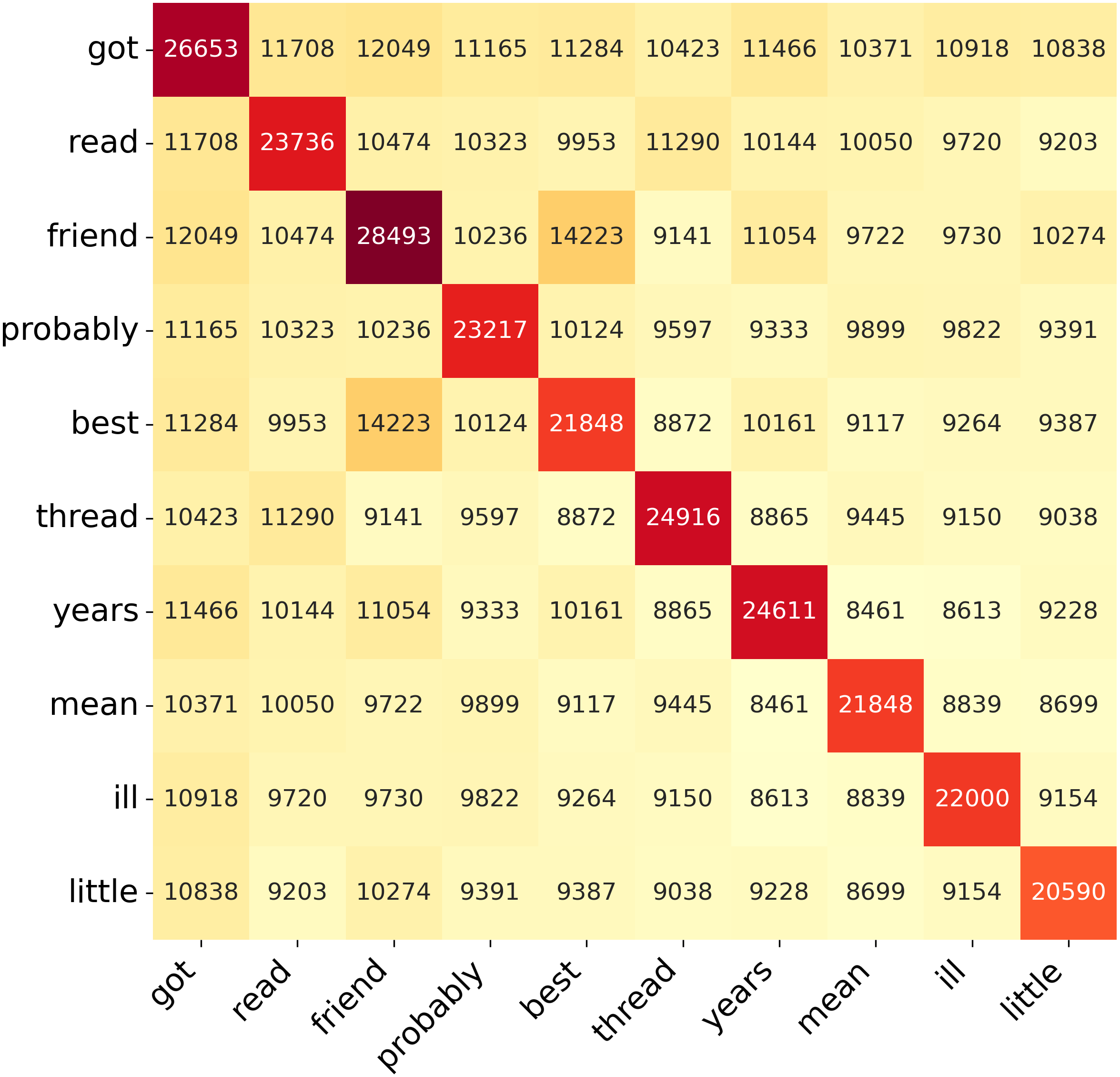}%
    \hspace{0.3cm}
    \includegraphics[scale=0.087]{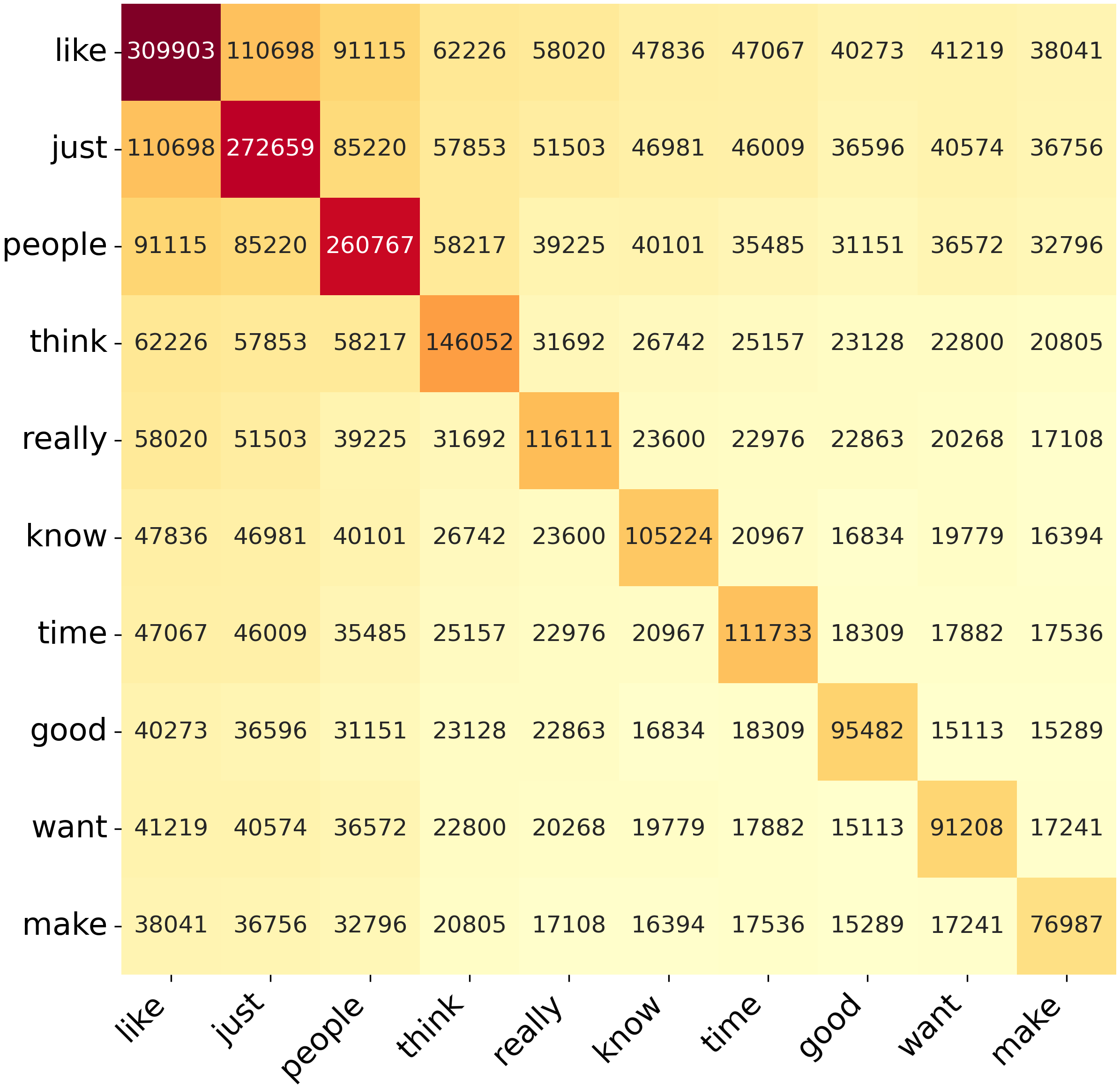}
    \caption{The visualization shows the co-occurrence matrix of the top 10 most frequent words in the real texts from the Kaggle (left) and Pandora (right) datasets.}
    \label{fig:occurrence}
    \vspace{-0.15in}
\end{figure}

A central issue highlighted in this work is that a key limitation shared by existing methods lies in their difficulty handling complex reasoning and multi-round language style modelling \citep{shahnazari2025you}. Personality detection goes beyond surface-level semantic analysis; it often relies on context-dependent psycholinguistic cues. For example, as shown in Figure~\ref{fig:occurrence}, words like “best” may carry limited explanatory power when viewed in isolation. However, analysing their contextual associations, such as interactions with surrounding words, co-occurrence patterns, and distributional structures, can reveal deeper personality-related signals. This is because users’ lexical choices, expression styles, and affective tones are typically distributed across multiple posts \citep{shen2025ll4g}, making it challenging for conventional models to accurately interpret how context influences the expression of personality traits. This observation forms the core intuition behind our design: effective personality modelling requires structured, hierarchical processing mechanisms that can dynamically identify, select, and integrate relevant contextual information dispersed throughout a user's posts.

However, personality datasets often exhibit severe class imbalance, as shown in Table~\ref{tab:data-statistics}. This is typically due to two fundamental factors. First, certain personality types are inherently less common in the general population, leading to an uneven distribution of labelled data. For example, types such as INFJ and INTJ account for less than 2\% of the population \citep{amirhosseini2020machine}, resulting in their significant under-representation. Second, different personality types vary in their levels of social media activity and self-disclosure \citep{teng2022understanding}. Extroverted individuals, for instance, are more likely to post frequently and openly, while introverted users may post less often or express themselves more subtly. As a result, extroverts tend to be over-represented in collected datasets, further amplifying the original distributional imbalance. In addition, social media posts are typically short, which exacerbates the challenge of extracting sufficient features for accurate modelling.

\begin{table}[htbp]
\caption{Kaggle and Pandora datasets information in terms of the class distribution.}
\centering
\scalebox{0.8}{
\begin{tabular}{c|c|c|c|c|c|c}
    \hline
     & \multicolumn{3}{c|}{\textbf{Kaggle}} & \multicolumn{3}{c}{\textbf{Pandora}}   \\
    \hline
     \textbf{Types} & \textbf{Train} & \textbf{Validation} & \textbf{Test} & \textbf{Train} & \textbf{Validation} & \textbf{Test} \\
    \hline
     I/E & 1194 / 4011 & 409 / 1326 & 396 / 1339 & 1162 / 4278 & 386 / 1427 & 377 / 1437 \\
     S/N & 610 / 4478 & 222 / 1513 & 248 / 1487 & 727 / 4830 & 208 / 1605 & 210 / 1604 \\
     T/F & 2410 / 2795 & 791 / 944 & 780 / 955 & 3549 / 1891 & 1120 / 693 & 1182 / 632 \\
     P/J & 2109 / 3096 & 672 / 1063 & 653 / 1082 & 2229 / 3211 & 770 / 1043 & 758 / 1056 \\
     Posts & 246794 & 82642 & 82152 & 523534 & 173005 & 174080 \\
    \hline
\end{tabular}
}
\label{tab:data-statistics}
\end{table}

Cognitive science offers valuable insights into similar problems. Human social cognition, such as personality perception, is a highly complex process involving hierarchical information processing in the brain, integrating both deliberate reasoning and intuitive pattern recognition. The cerebral cortex plays a pivotal role in deep cognitive processing \citep{zhang2025brain}, enabling individuals to comprehend intricate social contexts, infer social attributes, and construct long-term behavioural patterns. This global reasoning capability allows humans to integrate diverse information sources when forming judgments. However, cognition is not solely dependent on higher-order reasoning. The basal ganglia, together with their associated neural circuits, contribute to fast, automatic pattern recognition \citep{maalej2024comparative}, allowing the brain to detect familiar cues and make rapid, experience-based classifications. Within these circuits, dopaminergic signals play a crucial role by encoding prediction errors and reinforcing adaptive responses \citep{lerner2021dopamine}, thereby modulating learning and decision-making. Another essential component of social cognition is working memory, supported by the prefrontal cortex \citep{shirdel2024exploring}. For example, when reading a lengthy article, individuals do not retain every word but extract key information; similarly, during conversation, attention is directed toward the main content rather than peripheral details. Working memory enables the selective maintenance and manipulation of task-relevant information, supporting flexible and goal-directed behaviour. This coordinated interplay among different brain regions ensures the efficient allocation of cognitive resources \citep{liao2024cognitive} and underlies the adaptability of human social perception. 

In this paper, we present a \textbf{h}ierarchical \textbf{i}nformation \textbf{p}rocessing framework for \textbf{p}ersonality \textbf{d}etection, HIPPD, inspired by human cognitive mechanisms. Specifically, HIPPD leverages LLMs to emulate the deep cognition of the cerebral cortex, enabling comprehensive semantic understanding and capturing long-range dependencies within textual data. However, as in human cognition, not all extracted information is equally relevant \citep{shen2025less}. To address this, we introduce a dynamic memory module inspired by the prefrontal cortex, functioning as a working memory system. This module retains essential task-related features and dynamically updates memory content, selectively prioritising the most relevant information. The refined memory representations are then processed by a specialised set of lightweight models that mimic the function of the basal ganglia. A strict winner-takes-all selection mechanism is employed to dynamically route each input through only the most suitable specialised model, ensuring efficient and focused pattern recognition for personality-related traits. Furthermore, dopaminergic-like prediction error signals are employed to adaptively regulate both memory updating and the model selection process, reinforcing effective information processing and enhancing the system’s overall adaptability.

\vspace{0.05in}

\noindent \textbf{Main Contributions:} We propose HIPPD, a hierarchical framework for personality detection that draws inspiration from the human brain’s architecture. HIPPD integrates a global semantic encoder, a dynamic working memory module, and a specialist model routing layer with a strict winner-takes-all strategy. It uniquely incorporates prediction error feedback to adaptively regulate memory and routing, enabling robust, context-sensitive information processing. These mechanisms together address challenges such as class imbalance and short text, and generalize well to other feature-scarce tasks. Extensive experiments on two public benchmarks show that HIPPD consistently outperforms state-of-the-art baselines.

\section{Related Work}
\noindent \textbf{Deep Learning Methods:} Transformer and LSTM architectures have demonstrated strong performance in personality recognition tasks. For example, a study \citep{naz2024ai} focused on the identification of extraversion-related features by combining BERT and BiLSTM. Another approach introduced a label prompting method to enhance a language model's understanding of specific personality traits \citep{chen2023mining}. In addition, researchers have explored combining semantic features with word embedding techniques \citep{serrano2024combining}. Other work fine-tuned RoBERTa on unlabeled Telegram conversations, demonstrating its transferability in unstructured settings \citep{shahnazari2025you}. Furthermore, some studies have utilized knowledge graph techniques to enhance the model's capability in personality recognition \citep{ramezani2022text, zhu2022contrastive}.

\noindent \textbf{Data-Efficient Learning Methods:} Efficient and generalisable learning strategies are becoming a focus of this topic. For example, multi-task learning \citep{shen2025emoperso} has been shown to capture both emotion and personality signals in a shared representation space for collaborative training, effectively mitigating the problems of feature redundancy \citep{li2022multitask}. Researchers have used data rebalancing strategies to increase the weight of minority class samples, combined dynamic balancing strategies with difficult example mining \citep{jiang2024towards}. Federated learning has also been applied to personality recognition, where privacy-preserving mechanisms such as sentiment and topic preference modelling help enhance performance under distributed data settings without access to raw data \citep{zhao2023federated}. Furthermore, generative models combined with active learning can synthesise or select representative samples without increasing the annotation cost \citep{meng2024deep}.

\vspace{0.05in}

\noindent Unlike previous approaches that primarily rely on pre-trained models and uniform feature extraction, HIPPD introduces a hierarchical, brain-inspired framework that combines global reasoning, dynamic memory, and winner-takes-all pattern selection. This enables more effective identification and integration of salient personality cues across diverse and context-dependent user data.

\section{Our Novel HIPPD Model}
\begin{figure*}[htbp]
    \centering
    \includegraphics[scale=0.170]{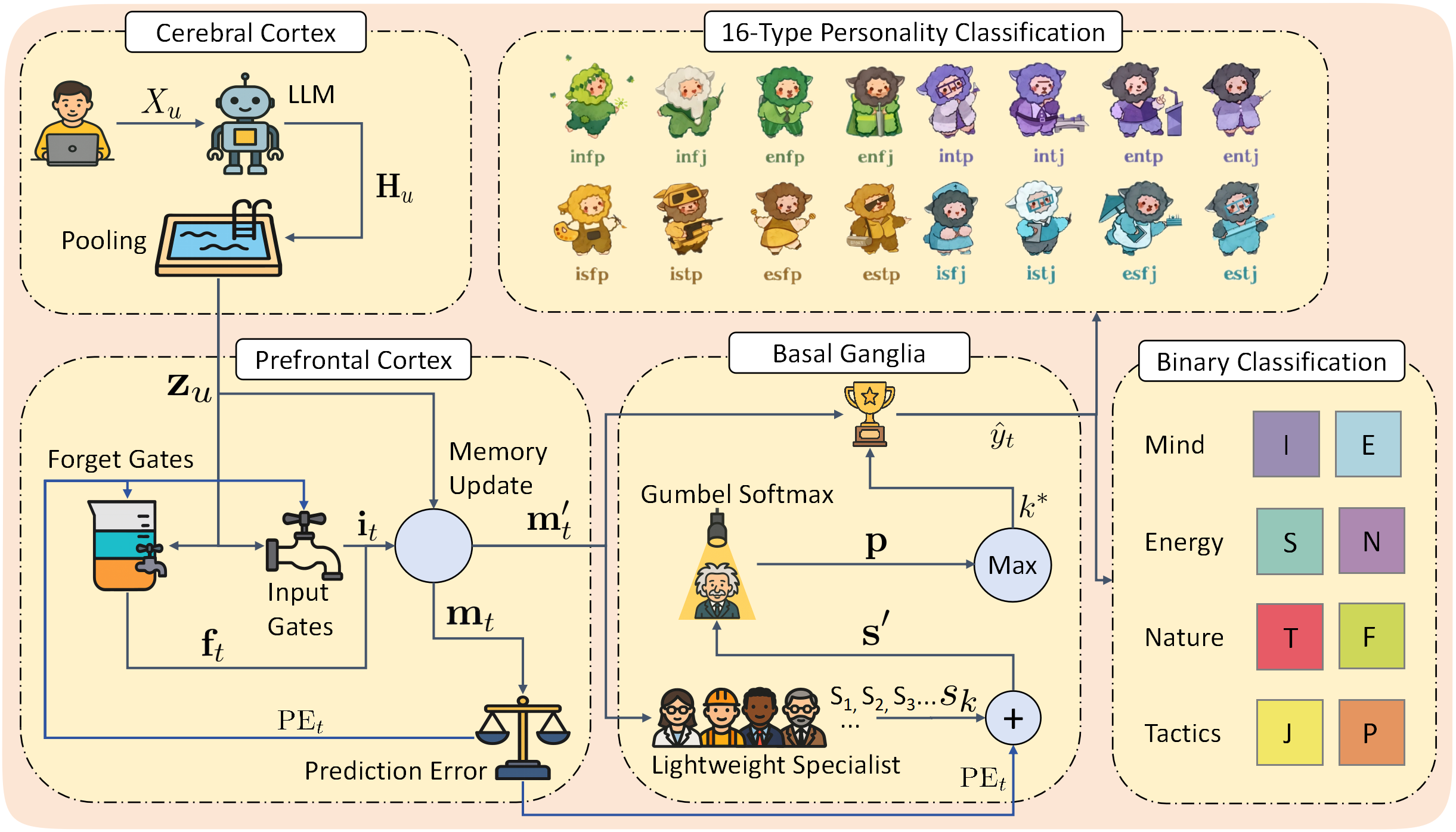} %
    \caption {Overview of the HIPPD architecture. The model simulates three core brain modules: the cerebral cortex, the prefrontal cortex, and the basal ganglia. Dopaminergic modulation adaptively regulates working memory and specialist selection.}
    \label{fig:diagram}
    \vspace{-0.20in}
\end{figure*}

HIPPD is a brain-inspired hierarchical model for personality detection from text. It comprises three main modules (Figure~\ref{fig:diagram}): a global semantic encoder simulating the cerebral cortex for high-level contextual abstraction; a dynamic working memory module inspired by the prefrontal cortex for selective feature retention and updating; and a parallel pool of lightweight specialist models mimicking the basal ganglia, where a strict winner-take-all routing mechanism activates only the most suitable specialist. Unified dopaminergic feedback adaptively regulates both memory gating and model selection, forming a closed-loop, self-optimising information processing system capable of robust operation across diverse contexts.

\subsection{Global Semantic Encoder}

Let $\mathcal{X} = \{x_1, x_2, \ldots, x_N\}$ denote the sequence of user-generated text posts, where $x_i$ represents the $i$-th post in the corpus. We employ an LLM, denoted as $f_{\theta}$ and parameterised by $\theta$, to encode each input sequence into a high-dimensional semantic space. Specifically, for each user, the model receives the concatenated text sequence $X_u = [x_1^{(u)}, \cdots, x_{M_u}^{(u)}]$, where $M_u$ is the number of posts for user $u$, allowing the model to compute feature dependencies within a broader contextual window.

The LLM backbone maps $X_u$ into a contextualised feature matrix $\mathbf{H}_u = f_{\theta}(X_u) \in \mathbb{R}^{M_u \times d}$, where $d$ is the hidden dimension and $\mathbf{H}_u = [\mathbf{h}_1, \cdots, \mathbf{h}_{M_u}]$ contains the contextual embeddings of each post. This step simulates the integrative and abstraction function of the cerebral cortex\footnote{The cerebral cortex is the highly folded outer layer of the brain, responsible for complex cognitive processes such as perception, language, reasoning, and conscious thought.}, enabling the model to capture long-range dependencies and nuanced language patterns across multiple posts.
To obtain a global semantic representation for the user, we aggregate the contextual features via a pooling function $\mathcal{P}$ (e.g., mean or attention-based pooling), i.e.,
$\mathbf{z}_u = \mathcal{P}(\mathbf{H}_u) \in \mathbb{R}^{d}$,
where $\mathbf{z}_u$ encodes the user's overall linguistic and contextual information.
This global representation $\mathbf{z}_u$ serves as the input to downstream modules, supporting dynamic memory operations and specialist pattern recognition in subsequent stages.

\subsection{Dynamic Working Memory Module}

Given the global semantic representation $\mathbf{z}_u$ obtained from the LLM, the dynamic working memory module is designed to selectively retain, update, and filter information in a task-adaptive manner, mimicking the working memory\footnote{Working memory refers to the brain's capacity to temporarily store and manipulate information necessary for complex tasks like reasoning, comprehension, and decision-making. It is primarily supported by the prefrontal cortex and is essential for adaptive, goal-directed behaviour.
} and gating functions of the prefrontal cortex\footnote{The prefrontal cortex is the foremost region of the frontal lobes, known for its central role in executive functions, including planning, decision-making, attentional control, and moderating social behaviour.}.
A central mechanism enabling the prefrontal cortex to perform flexible and adaptive control is its use of prediction error (PE) signals. Substantial neuroscientific and computational evidence \citep{schultz1997neural, lerner2021dopamine} demonstrates that dopaminergic neurons encode reward prediction error, which the prefrontal cortex exploits to drive synaptic plasticity, optimise working memory retention, and allocate cognitive resources to task-relevant information. This predictive modulation allows the brain to strengthen or suppress information based on feedback, supporting continual adaptation in dynamic environments. Motivated by this principle, we use prediction error as an analogue for dopaminergic modulation to regulate the gating dynamics of our working memory layer.

The memory state at time step $t$ is denoted as $\mathbf{m}_t \in \mathbb{R}^d$. For each new input $\mathbf{z}_u$, the memory layer computes gated updates as:
$\mathbf{i}_t = \sigma(\mathbf{W}_i \mathbf{z}_u + \mathbf{U}_i \mathbf{m}_{t-1} + \mathbf{b}_i),
\mathbf{f}_t = \sigma(\mathbf{W}_f \mathbf{z}_u + \mathbf{U}_f \mathbf{m}_{t-1} + \mathbf{b}_f)$,
where $\mathbf{i}_t$ and $\mathbf{f}_t$ are the input and forget gates, $\sigma(\cdot)$ is the sigmoid activation, and $\mathbf{W}_i, \mathbf{W}_f, \mathbf{U}_i, \mathbf{U}_f, \mathbf{b}_i, \mathbf{b}_f$ are trainable parameters.
The memory update equation is defined as:
\begin{equation}
\mathbf{m}_t = \mathbf{f}_t \odot \mathbf{m}_{t-1} + \mathbf{i}_t \odot \tanh(\mathbf{W}_c \mathbf{z}_u + \mathbf{U}_c \mathbf{m}_{t-1} + \mathbf{b}_c),
\end{equation}
where $\odot$ denotes element-wise multiplication, and $\mathbf{W}_c, \mathbf{U}_c, \mathbf{b}_c$ are learnable parameters.
To introduce adaptive regulation inspired by dopaminergic prediction error, we compute the PE signal as
$\mathrm{PE}_t = \ell(\hat{y}_t, y_t)$,
where $\hat{y}_t$ is the model's predicted label at time $t$, $y_t$ is the ground truth, and $\ell(\cdot,\cdot)$ is a suitable loss function (e.g., cross-entropy).
The PE signal modulates the gating parameters by
$\mathbf{i}_t' = \mathbf{i}_t + \alpha\, \mathrm{PE}_t,
\mathbf{f}_t' = \mathbf{f}_t - \beta\, \mathrm{PE}_t$,
where $\alpha, \beta > 0$ are modulation coefficients. This adaptive adjustment allows the memory layer to reinforce critical features when the error is high, and selectively forget less relevant information, achieving continual memory optimisation.
Finally, the updated memory state $\mathbf{m}_t'$ (using $\mathbf{i}_t', \mathbf{f}_t'$) is passed to the downstream specialist model pool for further processing and personality trait inference.

\subsection{Specialist Model Routing}

Biologically inspired by the parallel and competitive action selection circuits of the basal ganglia \footnote{The basal ganglia is a group of interconnected brain nuclei responsible for the selection, initiation, and regulation of actions and behaviours, playing a key role in both motor and cognitive decision-making.} We model this stage as a set of $K$ lightweight specialist models. Following the adaptive working memory update, the processed feature vector $\mathbf{m}_t' \in \mathbb{R}^d$ is fed into these specialist models, denoted $\{\mathcal{S}_1, \ldots, \mathcal{S}_K\}$. Each specialist model $\mathcal{S}_k$ is parameterised by $\phi_k$ and designed to recognise distinct personality-related patterns.

For a given input, a gating network computes a suitability score for each specialist, i.e.,
$s_k = \mathbf{w}_k^\top \mathbf{m}_t' + b_k, k=1, \ldots, K$,
where $\mathbf{w}_k \in \mathbb{R}^d$ and $b_k \in \mathbb{R}$ are gating parameters.
During training, a differentiable winner-takes-all routing is implemented via the Gumbel-Softmax estimator \citep{jang2016categorical}, i.e., 
$\mathbf{p} = \mathrm{Gumbel\text{-}Softmax}(\mathbf{s}; \tau)$,
where $\mathbf{s} = [s_1, \ldots, s_K]^\top$ and $\tau$ is the temperature parameter. During inference, a strict argmax is applied, i.e., $k^* = \arg\max_{k} s_k$,
so that only the most suitable specialist is activated for the final prediction. 
We adopt this strict winner-takes-all mechanism because it closely mirrors the action selection dynamics of the biological basal ganglia, which resolve competition among parallel neural circuits by allowing only the most salient pathway to be expressed while suppressing alternatives. This competition-driven selection is widely recognised as a key principle for efficient and non-redundant decision making in both neuroscience and cognitive science \citep{frank2001interactions, redgrave1999basal}.
The selected specialist receives $\mathbf{m}_t'$ as input and produces the output $
\hat{y}_t = \mathcal{S}_{k^*}(\mathbf{m}_t'; \phi_{k^*})$.
To maintain biological plausibility and enable continual specialisation, the PE signal computed in the previous stage is also used to adaptively modulate the gating network, i.e., $\mathbf{s}' = \mathbf{s} + \boldsymbol{\eta} \cdot \mathrm{PE}_t$, where $\boldsymbol{\eta} \in \mathbb{R}^K$ is a modulation vector. This feedback mechanism encourages exploration when errors are high and reinforces successful routing decisions when errors are low, mirroring dopaminergic action selection observed in the basal ganglia.

By combining strict competition (winner-takes-all), dynamic gating, and error-driven adaptation, this module enables efficient and adaptive pattern selection for robust personality trait inference.

\subsection{Personality Classification}

HIPPD supports both binary classification along each MBTI dimension and full 16-type personality classification.

Let $\hat{y}_t$ denote the output logit from the selected specialist model for input at time $t$. For each MBTI dimension (Introversion/Extraversion, Sensing/Intuition, Thinking/Feeling, Judging/Perceiving), a separate sigmoid-based classification head is applied, i.e.,
\begin{equation}
\hat{p}_{t}^{(d)} = \sigma(\mathbf{w}_{\mathrm{bin}}^{(d)\top} \hat{y}_t + b_{\mathrm{bin}}^{(d)}), \qquad d \in \{1,2,3,4\}
\end{equation}
where $\mathbf{w}_{\mathrm{bin}}^{(d)}$ and $b_{\mathrm{bin}}^{(d)}$ are the weights and bias for the $d$-th dimension, and $\sigma(\cdot)$ is the sigmoid function. The prediction for each dimension is given by $
y_{\mathrm{bin}}^{(d)} = \mathbb{I}[\hat{p}_{t}^{(d)} > 0.5]$, where $\mathbb{I}$ is the indicator function.
For joint prediction of the full MBTI type, we introduce a softmax-based multi-class head, i.e.,
$\hat{\mathbf{p}}_t^{(16)} = \mathrm{softmax}(\mathbf{W}_{16} \hat{y}_t + \mathbf{b}_{16})$, 
where $\mathbf{W}_{16} \in \mathbb{R}^{16 \times d}$ and $\mathbf{b}_{16} \in \mathbb{R}^{16}$ are the weight matrix and bias vector. The final predicted personality type is
\begin{equation}
y_{16} = \arg\max_{k \in \{1, \ldots, 16\}} \hat{\mathbf{p}}_t^{(16)}[k].
\end{equation}

During training, the total loss is computed as the sum of the binary cross-entropy loss for each dimension and the categorical cross-entropy loss for the 16-type prediction, enabling joint optimisation in both binary and multiclass settings.

\section{Experiments and Discussions}
\subsection{Experimental Setup}
\noindent \textbf{Datasets:} To ensure a fair comparison with previous work \citep{yang2023orders, hu2024llm, zhu2024integrating}, we selected the same two datasets, i.e., Kaggle\footnote{\url{https://www.kaggle.com/datasnaek/mbti-type}} and Pandora\footnote{\url{https://psy.takelab.fer.hr/datasets/all}}. The Kaggle MBTI dataset consists of data from approximately 8,675 users collected via the PersonalityCafe forum. Each user contributed up to 50 posts along with their self-reported MBTI labels. The posts are typically short (averaging about 26 words per post), cover a wide range of topics, and reflect diverse writing styles. In total, the dataset comprises around 430,000 individual text samples. The Pandora dataset is another large-scale benchmark constructed from Reddit. It contains about 9,067 users, each with dozens to hundreds of posts. MBTI labels are assigned by extracting type mentions from users’ self-introduction posts. Compared to the Kaggle dataset, Pandora offers greater linguistic diversity and contextual depth, as Reddit users tend to participate in a broader range of communities and topics. 
Both datasets are regarded as standard benchmarks for evaluating new methods in text-based MBTI personality computing.

\noindent \textbf{Implementation Details:} We adopt Qwen3-14B \citep{yang2025qwen3} as the frozen global semantic encoder. To align input lengths across Kaggle and Pandora, each user sequence is standardised to the median of 2048 tokens. The working-memory module is a single-layer gated memory network: input and forget gates are MLPs with sigmoid activations and width 4096 to match the LLM embedding; parameters use Xavier initialisation; state updates use $\tanh$. Positional modulation uses a coefficient of 0.1, with the positional signal normalised and clipped to $[-1,1]$ for stability. A memory dropout of 0.2 is applied after each update, and the updated memory vector is optionally linearly projected to match downstream input sizes. The specialist pool comprises five parallel lightweight models (CNN, LSTM, GCN, SVM, XGBoost); the CNN has 128 channels; the LSTM uses a hidden size of 512 with dropout 0.2; XGBoost and SVM use the same features but are trained post hoc; no parameters are shared across models. Training employs Gumbel-Softmax routing with temperature linearly decayed from 1.0 to 0.1 over the first 20 epochs; inference uses strict argmax for winner-take-all selection. The prediction error is the cross-entropy between predicted and true labels at each step, normalised to $[0,1]$ by mini-batch min–max rescaling; high error flattens gating logits to encourage exploration, while low error sharpens them to reinforce consistent selection. The modulation step size is 0.1. Experiments run on three NVIDIA H20 GPUs using the Adam optimiser with a learning rate of $1\times10^{-4}$. To prevent information leakage, tokens directly matching personality labels are removed during preprocessing. Data splits are 60\%/20\%/20\% for training/validation/test, and results are averaged over ten runs.

\noindent \textbf{Evaluation Metrics:} We employ the Macro-F1 as our metric, which has been utilised in previous studies \citep{yang2023orders, hu2024llm, zhu2024integrating}. Additionally, we used accuracy (ACC), precision (P), and recall (R) in various evaluation scenarios to provide a more comprehensive assessment.

\noindent \textbf{Comparative Models:} We first selected several popular classification baselines, including SVM \citep{cui2017survey}, XGBoost \citep{tadesse2018personality}, LSTM \citep{tandera2017personality}, and BERT \citep{keh2019myers}. We then evaluated the zero-shot and chain-of-thought (CoT) capabilities of advanced LLMs, including GPT-4o \citep{achiam2023gpt} and DeepSeek-V3 \citep{liu2024deepseek}. We also incorporated diverse fine-tuned architectures. AttRCNN \citep{xue2018deep} integrates attention into an RCNN structure with a CNN-Inception module for robust feature extraction. SN+Attn \citep{lynn2020hierarchical} uses a Sequence Network with dual attention at message and word levels to enhance signal relevance. Transformer-MD \citep{yang2021multi} employs Transformer-XL and memory mechanisms for disorder-agnostic post-integration with dimension-specific attention. PQ-Net \citep{yang2021learning} fuses psychological questionnaires and user text via cross-attention to capture personality cues. TrigNet \citep{yang2021psycholinguistic} constructs a heterogeneous tripartite graph with flow graph attention (GAT) for psycholinguistic integration. 
D-DGCN \citep{yang2023orders} dynamically builds graph structures, integrating multiple posts disorder-agnostically. DEN \citep{zhu2024enhancing} models long-term personality traits with GCN, short-term states with BERT, and enhances both via bidirectional interaction. MvP \citep{zhu2024integrating} introduces a multi-view mixture-of-experts (MoE) architecture that automatically models and integrates user posts from multiple perspectives. 
TAE \citep{hu2024llm} leverages LLM-generated text augmentation and label explanations, applying contrastive learning to improve psycholinguistic representation.

\subsection{Overall Results}
\noindent \textbf{Macro F1:} As shown in Table~\ref{tab:macro-f1}, our HIPPD achieves an average Macro-F1 score of 78.97\% on the Kaggle dataset, and 68.98\% on the Pandora dataset, surpassing the previous state-of-the-art model TAE by 6.90\% and 5.93\%, respectively. In addition, HIPPD consistently delivers substantial improvements on each individual dimension, demonstrating robust and comprehensive advantages across all evaluation aspects.

Specifically, compared with both meticulously tuned traditional machine learning models, such as SVM, XGBoost, and LSTM, as well as deep learning architectures incorporating complex modules like CNNs, GCNs and Transformers, HIPPD exhibits significant performance gains. Moreover, even when faced with LLMs renowned for their strong language understanding capabilities, such as GPT-4o and DeepSeek-V3, HIPPD maintains superior competitiveness. Notably, introducing CoT reasoning into LLMs does not bring additional benefits; on the contrary, it slightly degrades performance on this task. This observation suggests that the current general-purpose reasoning mechanisms in LLMs are unable to effectively identify and integrate the subtle personality cues distributed across multiple social media posts, and may even introduce irrelevant noise or hallucination.

\begin{table*}[htbp]
\caption{Comparison of HIPPD with state-of-the-art baselines in terms of the Macro-F1 (\%) scores across the four dimensions and their overall average (Avg).}
    \centering
    \scalebox{0.80}
    {
    \begin{tabular}{l|cccc|c|cccc|c}
        \hline
        \multirow{2}{*}{\textbf{Methods}} & \multicolumn{5}{c|}{\textbf{Kaggle}} & \multicolumn{5}{c}{\textbf{Pandora}} \\
        \cline{2-11}
        & I/E & S/N & T/F & P/J & \textbf{Avg} & I/E & S/N & T/F & P/J & \textbf{Avg} \\
        \hline
        SVM & 53.34 & 47.75 & 76.72 & 63.03 & 60.21 & 44.74 & 46.92 & 64.62 & 56.32 & 53.15 \\
        XGBoost & 56.67 & 52.85 & 75.42 & 65.94 & 62.72 & 45.99 & 48.93 & 63.51 & 55.55 & 53.50 \\
        LSTM & 57.82 & 57.87 & 69.97 & 57.01 & 60.67 & 48.01 & 52.01 & 63.48 & 56.21 & 54.91 \\
        BERT & 58.33 & 57.12 & 77.95 & 65.25 & 66.24 & 56.60 & 48.71 & 64.70 & 56.07 & 56.52 \\
        AttRCNN & 59.74 & 64.08 & 78.77 & 66.44 & 67.25 & 48.55 & 56.19 & 64.39 & 57.26 & 56.60 \\
        SN+Attn & 65.43 & 62.15 & 78.05 & 63.92 & 67.39 & 56.98 & 54.78 & 60.95 & 54.81 & 56.88 \\
        Transformer-MD & 66.08 & 69.10 & 79.19 & 67.50 & 70.47 & 55.26 & 58.77 & 69.26 & 60.90 & 61.05 \\
        PQ-Net & 68.94 & 67.65 & 79.12 & 69.57 & 71.32 & 57.07 & 55.26 & 65.64 & 58.74 & 59.18 \\
        TrigNet & 69.54 & 67.17 & 79.06 & 67.69 & 70.86 & 56.69 & 55.57 & 66.38 & 57.27 & 58.98 \\
        D-DGCN & 69.52 & 67.19 & 80.53 & 68.16 & 71.35 & 59.98 & 55.52 & 70.53 & 59.56 & 61.40 \\
        DEN & 69.95 & 66.39 & 80.65 & 69.02 & 71.50 & 60.86 & 57.74 & 71.64 & 59.17 & 62.35 \\
        MvP & 67.68 & 69.89 & 80.99 & 68.32 & 71.72 & 60.08 & 56.99 & 69.12 & 61.19 & 61.85 \\
        GPT-4o (Zero-shot) & 68.94 & 54.74 & 80.19 & 67.01 & 67.72 & 57.43 & 51.55 & 71.81 & 62.34 & 60.78 \\
        GPT-4o (CoT) & 66.71 & 61.92 & 77.31 & 60.88 & 66.71 & 61.02 & 57.23 & 66.48 & 57.00 & 60.43 \\
        DeepSeek-V3 (Zero-shot) & 69.76 & 58.61 & 75.71 & 64.37 & 67.11 & 61.39 & 54.31 & 68.05 & 58.17 & 60.48 \\
        DeepSeek-V3 (CoT) & 68.59 & 61.47 & 74.56 & 61.21 & 66.46 & 58.23 & 56.15 & 67.94 & 55.01 & 59.33 \\
        TAE & 70.90 & 66.21 & 81.17 & 70.20 & 72.07 & 62.57 & 61.01 & 69.28 & 59.34 & 63.05 \\
        \hline
        \textbf{HIPPD} & \textbf{73.28} & \textbf{77.84} & \textbf{85.17} & \textbf{79.60} & \textbf{78.97} & \textbf{65.98} & \textbf{63.27} & \textbf{77.69} & \textbf{68.97} & \textbf{68.98} \\
        \hline
    \end{tabular}
    }
    \label{tab:macro-f1}
\end{table*}

\noindent \textbf{Accuracy:} We further report accuracy as an auxiliary metric to comprehensively demonstrate the effectiveness of HIPPD. For consistency, we employ the similar set of baselines as in the Macro-F1 evaluation, including widely used text classification models as well as state-of-the-art architectures specifically designed for MBTI personality detection.

As shown in Table~\ref{tab:accuracy}, HIPPD achieves an average accuracy of 86.08\%, outperforming GPT-4o by 7.05\%. For the four specific MBTI dimensions, HIPPD again achieves substantial improvements across the board, and notably, it surpasses 90\% accuracy on S/N for the first time, and breaks through the 80\% threshold on both T/F and P/J. These results further validate the significant contributions of our proposed design to this task.

\begin{table*}[htbp]
    \hspace*{-0.01\textwidth}
    \begin{minipage}[t]{0.48\textwidth}
        \centering
        \caption{Comparison of HIPPD and state-of-the-art baselines on the Kaggle dataset in terms of accuracy.}
        \scalebox{0.70}{
        \begin{tabular}{lccccc}
            \hline
            \multicolumn{6}{c}{Kaggle} \\ \hline
            Method & I/E & S/N & T/F & P/J & \textbf{Avg} \\
            \hline
            SVM+TF-IDF & 71.00 & 79.50 & 75.00 & 61.50 & 71.75 \\
            XGBoost & 78.17 & 86.06 & 71.78 & 65.70 & 75.43 \\
            RNN & 67.66 & 62.08 & 77.86 & 63.75 & 67.84 \\
            LSTM+Glove & 72.58 & 80.53 & 74.06 & 62.59 & 72.44 \\
            LSTM+RMSprop & 77.42 & 86.36 & 72.94 & 66.33 & 75.76 \\
            BERT+MLP & 77.30 & 84.90 & 78.30 & 69.50 & 77.50 \\
            BERT+SVM & 79.06 & 86.04 & 74.29 & 65.42 & 76.20 \\
            BERT+CNN & 78.47 & 78.57 & 77.58 & 71.41 & 76.51 \\
            RoBERTa & 77.10 & 86.50 & 79.60 & 70.60 & 78.45 \\
            Transformer-MD & 76.75 & 86.47 & 78.29 & 68.05 & 77.39 \\
            TrigNet & 77.84 & 85.09 & 78.77 & 73.33 & 78.76 \\
            D-DGCN & 78.16 & 84.48 & 79.32 & 73.39 & 78.84 \\
            DeepSeek-V3 & 79.85 & 85.62 & 77.69 & 68.14 & 77.83 \\
            GPT-4o & 80.27 & 86.61 & 78.25 & 70.97 & 79.03 \\ \hline
            \textbf{HIPPD} & \textbf{85.41} & \textbf{91.99} & \textbf{85.34} & \textbf{81.59} & \textbf{86.08} \\
            \hline
        \end{tabular}
        }
        \label{tab:accuracy}
    \end{minipage}
    \hspace{0.18in}
    \begin{minipage}[t]{0.50\textwidth}
        \centering
        \caption{Performance comparison within the 16-type evaluation framework.}
        \scalebox{0.70}{
        \begin{tabular}{cccccc}
            \toprule
            Dataset & Method & Acc & P & R & F1 \\ 
            \midrule
            \multirow{8}{*}{Kaggle}
            & SVM & 21.44 & 11.29 & 10.92 & 11.08 \\
            & XGBoost & 23.17 & 11.61 & 9.89 & 10.73 \\
            & LSTM & 24.9 & 5.45 & 8.93 & 6.07 \\
            & BERT & 34.64 & 16.29 & 14.43 & 15.87 \\
            & D-DGCN & 40.58 & 27.30 & 23.28 & 24.23 \\
            & DeepSeek-V3 & 51.74 & 45.41 & 39.94 & 41.07 \\
            & GPT-4o & 54.12 & 49.84 & 42.35 & 45.16 \\
            & \textbf{HIPPD} & \textbf{73.02} & \textbf{86.77} & \textbf{61.44} & \textbf{69.03} \\
            \midrule
            \multirow{8}{*}{Pandora} 
            & SVM & 20.23 & 11.06 & 7.56 & 9.80 \\
            & XGBoost & 25.30 & 7.34 & 8.16 & 7.91 \\
            & LSTM & 23.65 & 2.69 & 6.04 & 3.39 \\
            & BERT & 27.40 & 5.03 & 8.49 & 5.65 \\
            & D-DGCN & 32.52 & 12.67 & 11.68 & 10.99 \\
            & DeepSeek-V3 & 38.70 & 46.01 & 25.36 & 30.91 \\
            & GPT-4o & 41.82 & 48.89 & 28.77 & 34.25 \\
            & \textbf{HIPPD} & \textbf{58.21} & \textbf{67.17} & \textbf{29.34} & \textbf{38.26} \\
            \bottomrule
        \end{tabular}
        }
        \label{tab:16class}
    \end{minipage}
\end{table*}

\noindent \textbf{16-type Evaluation:} We further adopt a novel 16-type evaluation framework to assess the generalization and adaptability of HIPPD across different practical scenarios. In this setting, the model is required to directly classify all 16 MBTI personality types. In contrast to the conventional four-dimensional binary classification, which often requires expert psychological knowledge, the 16-type framework offers a more intuitive and user-friendly experience for the general public. This approach also better aligns with the common preference for obtaining personalized “badges” rather than relying on complex psychological profiles.

As shown in Table~\ref{tab:16class}, we selected a range of advanced models that demonstrated strong performance in previous evaluations as baselines. We observed that under the 16-type setting, all models experienced a significant drop in performance, with some baseline scores even falling to single digits. This highlights the inherent difficulty of direct 16-type MBTI classification. These results emphasize that simply stacking features or increasing model capacity is insufficient to capture the subtle, context-dependent signals that underpin fine-grained personality expression in real-world social media. Notably, HIPPD maintains a significant performance advantage and exhibits a much smaller decline than all competing baselines. This points to a promising direction for future multi-class user modelling tasks. Specifically, mechanisms such as dynamic context selection, continual feedback-driven adaptation, and specialized pattern recognition should be incorporated, as these are especially critical for maintaining robustness and generalization under extreme class imbalance and distributed weak signals.

\begin{table*}[htbp]
    \caption{Ablation results of HIPPD with different
component configurations. The tests were conducted on Kaggle and Pandora datasets in terms of the evaluation metric Macro-F1 (\%) scores across the four dimensions and their average (Avg).}
    \centering
    \scalebox{0.66}
    {
    \begin{tabular}{l|cccc|c|cccc|c}
        \hline
        \multirow{2}{*}{\textbf{Components}} & \multicolumn{5}{c|}{\textbf{Kaggle}} & \multicolumn{5}{c}{\textbf{Pandora}} \\
        \cline{2-11}
        & I/E & S/N & T/F & P/J & \textbf{Avg} & I/E & S/N & T/F & P/J & \textbf{Avg} \\
        \hline
        Vallina Qwen3-14B & 64.75 & 62.31 & 71.40 & 62.62 & 65.27 & 56.82 & 52.62 & 65.73 & 55.70 & 57.72 \\
        Replace Qwen3 with BERT & 73.21 & 69.78 & 80.34 & 69.25 & 73.15 & 62.33 & 59.17 & 70.77 & 61.81 & 63.52 \\
        Replace Qwen3 with GPT-4o & \textbf{73.61} & 77.25 & 84.98 & 79.33 & 78.79 & 65.62 & 62.91 & 77.32 & 68.65 & 68.63 \\
        Replace Attention Pooling with Mean Pooling & 73.20 & 77.81 & 84.99 & 79.35 & 78.84 & 65.72 & 63.02 & 77.50 & 68.72 & 68.74 \\
        w/o Working Memory & 70.05 & 71.02 & 82.62 & 72.27 & 73.99 & 61.15 & 58.88 & 74.13 & 63.09 & 64.31 \\
        Replace Gated Memory with MLP & 71.39 & 71.58 & 82.95 & 74.29 & 75.05 & 62.24 & 58.03 & 74.81 & 64.30 & 64.85 \\
        w/o PE Signal & 72.40 & 74.66 & 83.11 & 77.03 & 76.80 & 63.28 & 60.69 & 76.40 & 65.93 & 66.58 \\
        Replace Winner-Takes-All with Soft Routing & 72.58 & 75.28 & 84.57 & 77.91 & 77.59 & 64.07 & 62.13 & 76.88 & 67.53 & 67.65 \\
        Replace Gating Network with Random Routing & 73.83 & 71.32 & 81.94 & 69.87 & 74.25 & 62.39 & 59.65 & 71.52 & 60.30 & 63.47 \\
        \hline
        \textbf{HIPPD} & 73.28 & \textbf{77.84} & \textbf{85.17} & \textbf{79.60} & \textbf{78.97} & \textbf{65.98} & \textbf{63.27} & \textbf{77.69} & \textbf{68.97} & \textbf{68.98} \\
        \hline
    \end{tabular}
    }
    \label{tab:ablation}
\end{table*}

\subsection{Ablation Study}

We conducted a detailed ablation study to confirm the contribution of each key component to our overall design, as shown in Table~\ref{tab:ablation}. 

\noindent \textbf{Global Semantic Encoder:} We first focused on the global semantic encoder, including the vanilla Qwen3-14B model, and versions in which Qwen3 is replaced by BERT or GPT-4o. The vanilla Qwen3-14B configuration does not contain any hierarchical or memory-based modules. Under this setting, the Macro-F1 scores drop substantially to 65.27\% on Kaggle and 57.72\% on Pandora, representing decreases of 13.70\% and 11.26\% respectively compared to the full HIPPD. This performance gap highlights that real-world social media data, which is characterized by more dispersed linguistic features and weaker personality signals, benefits substantially from structured information processing. When the backbone model is replaced with BERT, Macro-F1 decreases by 5.82\% on Kaggle and 5.46\% on Pandora. Although BERT remains a strong language model, its ability to capture long-range dependencies and subtle linguistic cues is somewhat inferior to advanced LLMs like Qwen3. This underscores the importance of high-capacity semantic modelling in complex and noisy environments. When Qwen3 is replaced by GPT-4o, the overall performance of both HIPPD variants becomes very close, with each achieving slightly better results on some individual dimensions. This result demonstrates that the HIPPD architecture is robust and generalizes well to different large language model backbones. The advantages brought by the carefully designed HIPPD outweigh the marginal differences between top-tier LLMs.

We further examined the impact of different pooling strategies on the global semantic encoder. Replacing attention pooling with mean pooling had only a negligible effect on performance, and the model still maintained very strong results. This similar outcome validates our design, which encodes rich contextual dependencies and global semantic signals, ensuring that the overall representation remains highly effective regardless of the pooling strategy employed.

\noindent \textbf{Dynamic Working Memory:} We next ablated the memory components in our model to evaluate their individual contributions. Removing the working memory module deprives HIPPD of the ability to highlight and update important features within user posts, which severely limits its capacity to aggregate distributed personality signals from fragmented data. When the gated memory mechanism is replaced with a simple MLP, performance also declines, although the impact is less severe than that of removing working memory entirely. This is because the memory structure remains, but the absence of gating eliminates the module’s ability to dynamically control which information is remembered or forgotten, thereby reducing the model’s flexibility in handling noisy or context-dependent signals.

We also examined the impact of the prediction error (PE) signal on the memory and routing modules. After removing the PE signal, the Macro-F1 drops to 76.80\% on Kaggle and 66.58\% on Pandora. This is because the PE signal enables the model to dynamically adjust memory updates and specialist routing based on recent prediction errors. Without this task-focused feedback mechanism, the model loses its ability for continual self-correction and adaptive information selection, which is especially detrimental on the more complex Pandora dataset.

\noindent \textbf{Specialist Model Routing:} To verify the importance of the strict winner-takes-all selection mechanism, we modified the routing strategy in two different ways. First, we implemented soft routing, in which all specialist models are combined using a softmax-weighted aggregation, instead of activating only the most suitable expert for each input as in the original design. This soft assignment weakens the competition among experts and allows every specialist to contribute partially to the final prediction. Second, we replaced the learned gating network with random routing. In this setting, for each input, one specialist model is assigned completely at random, without considering the suitability score or any data-driven criteria. Both ablation approaches lead to reduced model performance. The soft routing method causes a slight decrease in Macro-F1 (dropping to 77.59\% on Kaggle and 67.65\% on Pandora), indicating that allowing all specialist models to participate partially, rather than selecting only the most appropriate expert, reduces the clarity of pattern recognition. In contrast, random routing results in a much larger decline (dropping to 74.25\% on Kaggle and 63.47\% on Pandora), confirming that data-driven expert assignment is critical for robust and accurate personality detection, especially in the presence of noisy and heterogeneous social media data.

\section{Qualitative Analysis}

\begin{figure}[htbp]
    \centering
    \includegraphics[scale=0.06]{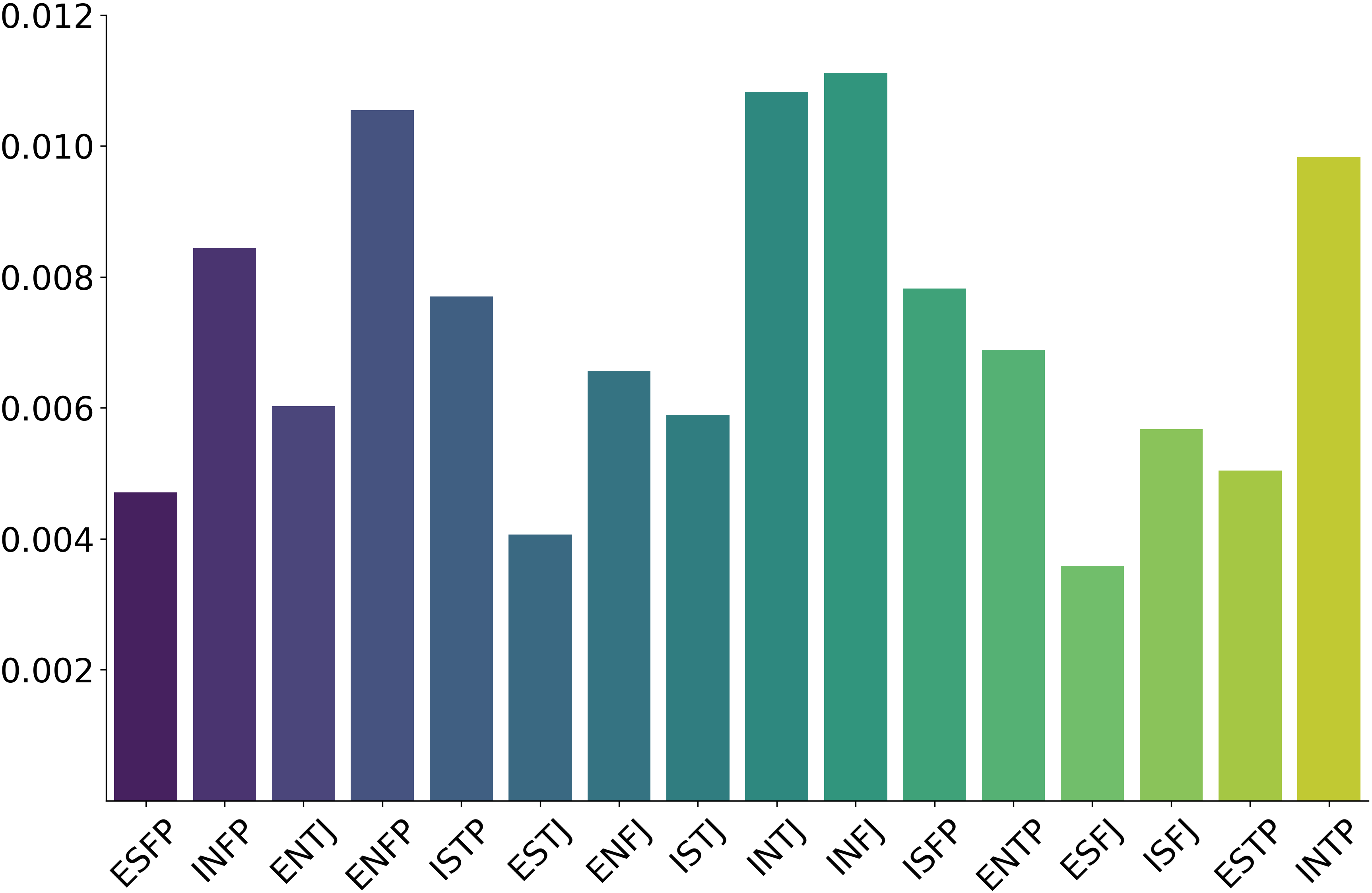}%
    \hspace{0.2cm}
    \includegraphics[scale=0.06]{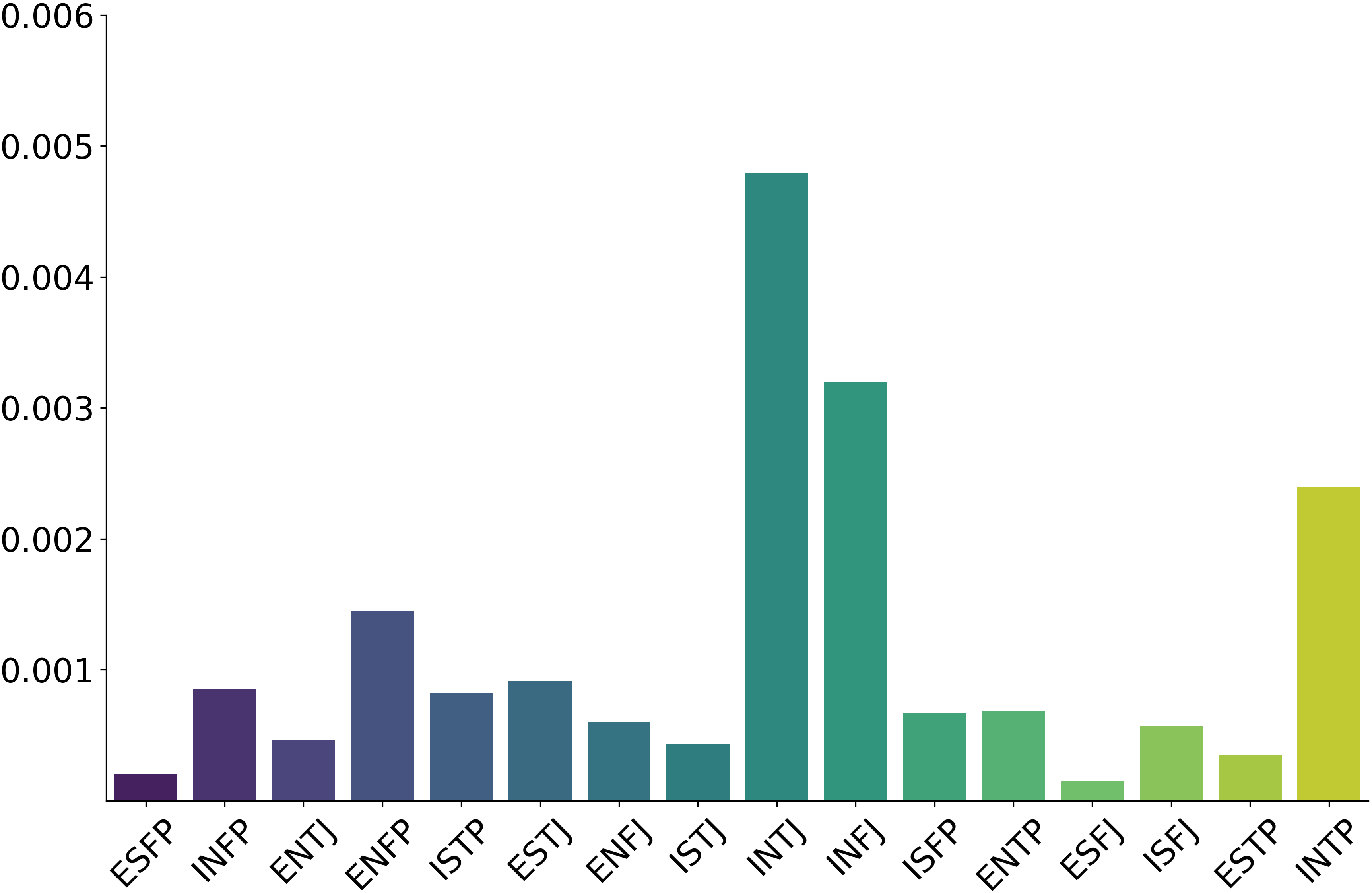}
    \caption{The visualization shows the mutual information between the top 10 most frequent words and all MBTI types in the Kaggle (left) and Pandora (right) datasets.}
    \label{fig:mutual}
\end{figure}

As shown in Figure~\ref{fig:mutual}, we use bar charts to visualise the mutual information (MI) between the ten most frequent words and all MBTI types in both the Kaggle (left) and Pandora (right) datasets. MI is a metric that measures the degree of dependency between two random variables; higher MI values indicate stronger associations and greater shared information. We leverage MI to assess the contribution of each frequent word to the MBTI types.

In the Kaggle dataset, although the INFJ type has the highest MI value, it is only around 0.012. The MI values for INTJ, ENFP, and other types decrease progressively, indicating that the associations between these personality types and the most frequent words are generally weak. In contrast, the Pandora dataset demonstrates overall lower MI values, with the peak remaining below 0.006 for any type. This reflects even weaker direct associations between frequent words and MBTI types in natural social media language.

These findings further highlight the need for hierarchical and structured modelling approaches, such as our HIPPD, which can move beyond surface-level word statistics to capture subtle, distributed, and context-dependent signals of personality expression. Moreover, the limited benefits of individual words or local features, especially in more diverse and ecologically valid datasets like Pandora, underscore the importance of integrating global dependencies, multi-level representations, and relational structures, as realised by HIPPD.

\section{Conclusion}

This paper proposes HIPPD, a brain-inspired hierarchical framework for personality detection. By modelling the division of labour among the cerebral cortex, prefrontal cortex, and basal ganglia, HIPPD achieves powerful global reasoning, dynamic memory selection, and context-sensitive pattern recognition. The unified dopaminergic feedback mechanism enables adaptive regulation of both working memory and specialist selection, forming a closed-loop, self-optimising system. Extensive experiments on two benchmark datasets demonstrate that HIPPD outperforms state-of-the-art baselines across multiple evaluation settings, while ablation studies confirm the critical contributions of all key components. In future work, we plan to extend this framework to other psychological and cognitive modelling tasks and to further explore continual adaptation in real-world dynamic environments.




\bibliography{main}






\end{document}